\documentclass[lettersize,journal]{IEEEtran}
\usepackage{amsmath,amsfonts,amssymb}
\usepackage{algorithmic}
\usepackage{algorithm}
\usepackage{array}
\usepackage[caption=false,font=normalsize,labelfont=sf,textfont=sf]{subfig}
\usepackage{textcomp}
\usepackage{stfloats}
\usepackage{url}
\usepackage{verbatim}
\usepackage{graphicx}
\usepackage{cite}
\usepackage{soul}
\usepackage{booktabs}
\usepackage{xcolor}
\usepackage{bbding}
\usepackage[misc]{ifsym}
\usepackage{colortbl}
\usepackage{multirow}
\usepackage{wrapfig}
\usepackage{caption}
\usepackage{tcolorbox}
\usepackage[pagebackref,breaklinks,colorlinks]{hyperref}

\usepackage[capitalize]{cleveref}

\usepackage[switch]{lineno}

\begin{document}
%\linenumbers

\title{Point Linguist Model: Segment Any Object via Bridged Large 3D-Language Model}

\author{Zhuoxu~Huang$^*$,
        Mingqi~Gao$^*$,
        and Jungong~Han\textsuperscript{\Envelope},~\IEEEmembership{Senior Member,~IEEE}
        % <-this % stops a space
\thanks{This work was supported in part by National Natural Science Foundation of China No.  62441235, and is also supported by Beijing Natural Science Foundation (L257005).}
\thanks{Zhuoxu Huang is with the Department of Computer Science, Aberystwyth University, Aberystwyth SY23 3DB, U.K. (e-mail: zhh6@aber.ac.uk).}
\thanks{Mingqi Gao is with the School of Computer Science, University of Sheffield, Sheffield S10 2TN, U.K. (e-mail: mingqi.gao@outlook.com).}
\thanks{Jungong Han is with the Department of Automation, Tsinghua University, Beijing, 100084, China. (e-mail: jungonghan77@gmail.com).}
\thanks{*Equal contributions.}
\thanks{\textsuperscript{\Envelope} Corresponding author Jungong Han}
}

% The paper headers
% \markboth{Journal of \LaTeX\ Class Files,~Vol.~14, No.~8, August~2021}%
% {Shell \MakeLowercase{\textit{et al.}}: A Sample Article Using IEEEtran.cls for IEEE Journals}

% \IEEEpubid{0000--0000/00\$00.00~\copyright~2021 IEEE}
% Remember, if you use this you must call \IEEEpubidadjcol in the second
% column for its text to clear the IEEEpubid mark.

\maketitle

\begin{abstract}
3D object segmentation with Large Language Models (LLMs) has become a prevailing paradigm due to its broad semantics, task flexibility, and strong generalization. However, this paradigm is hindered by representation misalignment: LLMs process high-level semantic tokens, whereas 3D point clouds convey only dense geometric structures. In prior methods, misalignment limits both input and output. At the input stage, dense point patches require heavy pre-alignment, weakening object-level semantics and confusing similar distractors. At the output stage, predictions depend only on dense features without explicit geometric cues, leading to a loss of fine-grained accuracy. 
To address these limitations, we present the Point Linguist Model (PLM), a general framework that bridges the representation gap between LLMs and dense 3D point clouds without requiring large-scale pre-alignment between 3D-text or 3D-images. 
Specifically, we introduce Object-centric Discriminative Representation (OcDR), which learns object-centric tokens that capture target semantics and scene relations under a hard negative-aware training objective. 
This mitigates the misalignment between LLM tokens and 3D points, enhances resilience to distractors, and facilitates semantic-level reasoning within LLMs.
For accurate segmentation, we introduce the Geometric Reactivation Decoder (GRD), which predicts masks by combining OcDR tokens carrying LLM-inferred geometry with corresponding dense features, preserving comprehensive dense features throughout the pipeline. Extensive experiments show that PLM achieves significant improvements of +7.3 mIoU on ScanNetv2 and +6.0 mIoU on Multi3DRefer for 3D referring segmentation, with consistent gains across 7 benchmarks spanning 4 different tasks, demonstrating the effectiveness of comprehensive object-centric reasoning for robust 3D understanding. Code is available at \href{https://github.com/zhh6425/PLM}{https://github.com/zhh6425/PLM}.
\end{abstract}


%%%%%%%%% BODY TEXT
\section{Introduction}
\label{sec:intro}

\IEEEPARstart{R}ecent advancements in Multi-modal Large Language Models (MLLMs) for point cloud understanding \cite{3urllm2025xiong, selfroi2025shi, weaklysupervised3dvlm2025xu, qi2025shapellm, xu2025pointllm, hong2023dllm, tang2024minigptd, qi2024gptpoint} have introduced a new interaction paradigm, enabling natural language-based communication with point cloud scenes. These pioneers have inspired efforts for the 3D object segmentation tasks \cite{dai2017scannet, rozenberszki2022scannet200, armeni2016s3dis, chen2020scanrefer, achlioptas2020referit3d, zhang2023multi3drefer} with LLMs. Despite the success in 2D MLLMs \cite{lai2024lisa, wang2024llmseg, rasheed2024glamm, xu2023ullava, yuan2025sa2va, gong2025devil, gong2025reinforcing, lin2025glus, ding2025multimodal}, without a segmentation priori like SAM \cite{kirillov2023sam}, successful paradigms like LISA \cite{lai2024lisa} cannot be replicated. 

The greatest obstacle is the representation misalignment between dense 3D points and discrete semantic tokens processed by pretrained LLMs, which affects the input and output in 3D object segmentation with LLMs. At input, it limits object-level semantics and renders localization unreliable under semantically similar distractors. Specifically, previous methods directly tokenize dense point patches like ViT \cite{dosovitskiy2020ViT} and strongly rely on large-scale pre-alignment between 3D-text or 3D-images \cite{he2025segpoint,deng20253dllava}.
Though effective in 3D object captioning or 3D question-answering \cite{xu2025pointllm, hong2023dllm}, such costs become impractical for complex fine-grained 3D understanding. More importantly, this patch-based tokenization isolates local geometry and ignores object boundaries and semantic relations. In cluttered scenes composed of complex objects, dense point patches inherently lack target-level structure and semantic cues about the object and its context. Consequently, LLMs struggle to distinguish the target from semantically similar distractors. At output, decoding relies solely on dense features and lacks object-aware geometric conditioning from the LLM. As a result, fine geometric cues are not preserved, yielding suboptimal dense predictions. 

\begin{figure*}[t]
    \centering
    \includegraphics[width=0.9\linewidth]{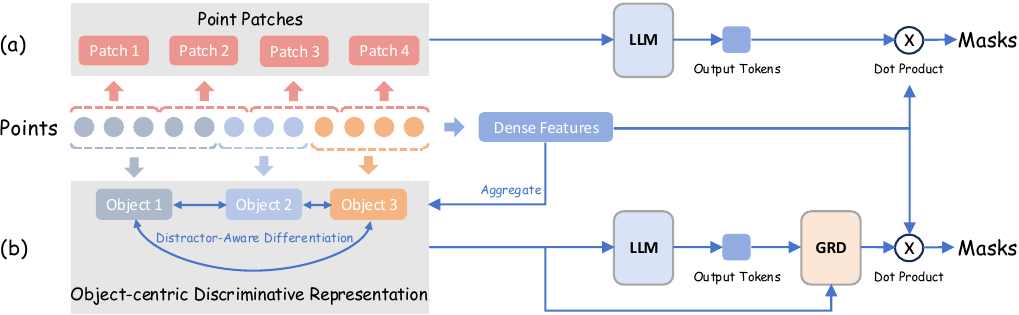}
    \caption{Comparison between (a) previous approaches and (b) our PLM. Given an input point cloud (colors denote different objects), previous approaches partition points into patches, ignoring object boundaries and target-level semantics. In contrast, PLM constructs an Object-centric Discriminative Representation (OcDR) from dense point features under distractor-aware supervision, capturing target-level semantics and explicitly assigning each token to a specific object while preserving inter-object differentiation. At output, previous approaches rely solely on dense scene features for the final prediction. To leverage geometric cues within the LLM reasoning pipeline, PLM injects dense features into the LLM and reactivates preserved details via the Geometric Reactivation Decoder (GRD).}
    \label{fig:architecture_comparison}
\vspace{-10pt}
\end{figure*}

\begin{figure}
    \centering
    \includegraphics[width=0.9\linewidth]{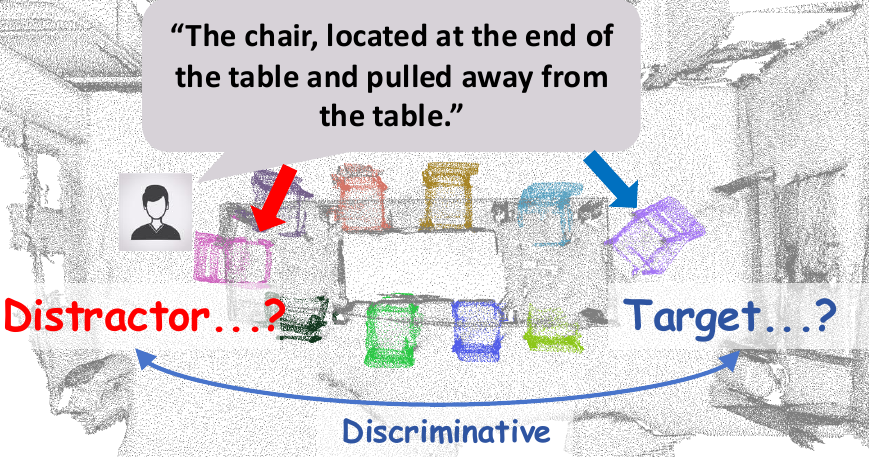}
    \caption{Example for distractor. In a complex scene where multiple objects are semantically related to the instruction (i.e., ``Chair"), only one is the ground truth.}
    \label{fig:example}
\vspace{-15pt}
\end{figure}

To address these issues, we propose \textbf{Point Linguist Model} (PLM) that bridges the representation gap between LLMs and dense 3D point clouds with our Object-centric Discriminative Representation (OcDR) and Geometric Reactivation Decoder (GRD). Firstly, our OcDR uses the object-centric (OC) tokens as the visual input of the LLM. As seen in the left panel of Figure~\ref{fig:architecture_comparison}, the OC token naturally maintains object-wise distinctions, allowing the LLM to directly access object-oriented features. The straightforward information injection helps the LLM to better identify different objects and their relationships in the scene. Secondly, based on the captured object relationships, we propose a distractor-supervised mechanism to refine object differentiation. As shown in Figure \ref{fig:example}, it selects hard negative distractors—objects with semantic proximity to the target—and incorporates them as additional supervision signals. By explicitly supervising the target with these distractors during training, our model enhances object identity discrimination, leading to more accurate and robust segmentation. For accurate segmentation, our GRD preserves comprehensive dense features throughout the pipeline. As shown in the right panel of Figure~\ref{fig:architecture_comparison} (b), we merge the comprehensive dense features into the OC tokens and preserve scene details during the LLM's reasoning pipeline. In the decoding process, the LLM-inferred geometry with corresponding dense features is reactivated through an attention mechanism before propagating to the final output mask.

In general, PLM integrates OcDR with GRD to systematically advance 3D object segmentation by aligning dense 3D geometry with LLM reasoning, informing point-level prediction with dense geometric cues. Specifically, OcDR generates discriminative object-centric representations via OC tokens with preserved dense geometric features and distractor-supervised learning, thereby enabling effective language understanding and enhanced object differentiation. Complementarily, GRD reactivates the preserved dense geometric features throughout the reasoning pipeline during decoding for accurate masks. Together, these two stages establish a coherent flow from dense 3D geometry to language reasoning and back to fine-grained segmentation outputs. We conduct comprehensive experiments across 7 datasets covering 4 downstream tasks, including open vocabulary instance segmentation (OVIS), open vocabulary semantic segmentation (OVSS), referring expression segmentation (RES), and generalized referring expression segmentation (GRES). In summary, the contributions of this work are threefold: 

\begin{itemize}
\item We identify the representation misalignment between dense 3D point clouds and LLMs, and propose Object-centric Discriminative Representation (OcDR) that uses object-centric tokens as an effective bridge representation, enabling structured object-level processing while preserving semantic relationships in 3D scenes.
\item We introduce a distractor-supervised learning mechanism within OcDR that leverages semantically similar hard negatives to enhance object discrimination, and develop the Geometric Reactivation Decoder (GRD) that maintains dense feature information throughout the LLM reasoning pipeline for precise segmentation.
\item We construct the Point Linguist Model, unifying the above innovations. With the language instruction, our model supports challenging open-vocabulary segmentation and referring segmentation tasks. Extensive experiments demonstrate the superiority of our approach.
\end{itemize}

\section{Related Works}
\label{sec:related}

\subsection{3D Object Segmentation}

In recent applications involving human language reasoning, new point cloud segmentation tasks have emerged, such as 3D referring segmentation \cite{chen2020scanrefer, achlioptas2020referit3d}, which aims to generate a segmentation mask from a natural language expression referring to a specific instance in the scene. A recent benchmark \cite{zhang2023multi3drefer} highlights the need for flexible segmentation of zero, one, or multiple objects from descriptive text, defining it as a generalized referring segmentation task \cite{wu20243dgres}. On the other hand, open-vocabulary semantic segmentation \cite{zhu2024opendiffusion, jiang2024openfoundation, wang2025xmask3d, peng2023openscene} requires segmenting objects based on class names from an open vocabulary, enabling recognition beyond predefined categories. Furthermore, open-vocabulary instance segmentation \cite{takmaz2023openmask3d, nguyen2024open3dis, boudjoghra2024openyolo} extends this concept by distinguishing individual object instances. Recently, the growing demand for intuitive human-machine interaction in robotics and embodied intelligence has driven interest in universal object segmentation, including the word-level 3D open-vocabulary and sentence-level 3D referring segmentation.

Previous expert models have notable limitations. For referring segmentation, those single-object segmentation methods \cite{huang2021textguided, wu2024dstmn, qian2024xrefsegd, he2024refmask3d} cannot meet the complexity of real-world applications. Some approaches \cite{zhang2023multi3drefer, wu20243dgres} have extended segmentation to multiple objects using multiple queries. However, they remain constrained to referring expressions and fail to support a diverse range of tasks under free-form language instructions. Recent advancements in 3D open-vocabulary segmentation have also led to the development of several notable methods. Most of them \cite{peng2023openscene, takmaz2023openmask3d, nguyen2024open3dis, boudjoghra2024openyolo} are heavily reliant on the fusion of image embeddings or well-pretrained 2D instance segmentors. OpenIns3D \cite{huang2024openins3d} presents a ``mask-snap-lookup" framework for 3D open-vocabulary instance segmentation, achieving state-of-the-art performance across datasets without relying on aligned 2D images. Our research investigates unifying both the 3D open-vocabulary segmentation and the 3D referring segmentation with an LLM architecture, achieving a significant advancement in both tasks.

\begin{figure*}[tbp]
    \centering
    \includegraphics[width=0.95\linewidth]{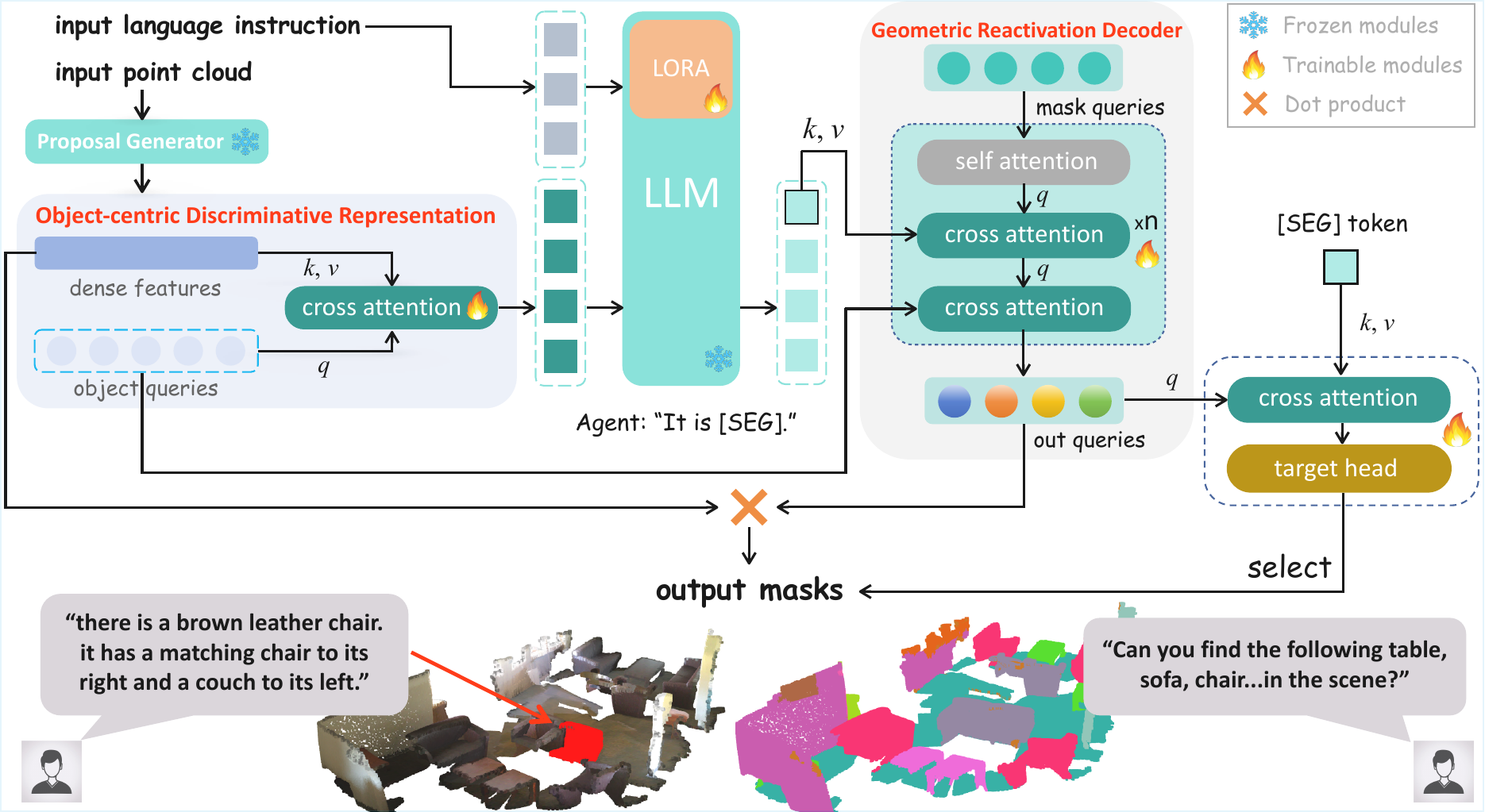}
    \caption{Overall architecture of the proposed Point Linguist Model. We propose OcDR to bridge the input pipeline from dense point cloud input to multi-modal LLM interaction, and design GRD to bridge the output pipeline from LLM outputs to dense segmentation. The proposed model can handle different tasks by adapting to different language instructions.}
    \label{fig:main_architecture}
\vspace{-15pt}
\end{figure*}

\subsection{Multi-modal Large Language Models}

Multi-modal Large Language Models (MLLMs) have garnered significant attention from researchers for their powerful ability to transfer the advanced reasoning capabilities of LLMs into the visual domain. Foundation models in both 2D \cite{liu2023llava, zhu2023minigpt, alayrac2022flamingo, tong2024cambrian} and 3D \cite{xu2025pointllm, hong2023dllm, liu2024unidllm, qi2024gptpoint} have focused on aligning image/point cloud and text features through instruction tuning, laying the foundation for vision Large Language Models. Recent approaches have increasingly focused on the more challenging task of dense, pixel-level prediction, e.g., LISA \cite{lai2024lisa} advances the 2D object segmentation by integrating the Segmentation Anything Model (SAM) \cite{kirillov2023sam} and LLaVA \cite{liu2023llava}. Such a paradigm shows that MLLMs can derive segmentation capacity from an a priori model like SAM. In the 3D domain, SegPoint \cite{he2025segpoint} also built a unified framework for various segmentation tasks, combining pre-trained Uni3D \cite{zhou2023uni3d} with the LLaMA model \cite{touvron2023llama2}. However, lacking a foundational segmentation model like SAM in the 3D domain, dense prediction requires MLLMs to learn scene parsing implicitly—a process that demands massive paired data \cite{mamaghan2024exploring, huang2023chatscene} and is therefore data-inefficient and suboptimal.

In this paper, we assert that such a common limitation is posed by the representation misalignment between the dense 3D point clouds and the object, which hampers the 3D object segmentation with complex spatial relationships between objects. Contrasting with existing models, our research investigates the OcDR as an effective and structured format to present the complex point cloud scene, further advancing the point cloud segmentation tasks.

\section{Proposed Method}
\subsection{Background and Overview}

We first give a brief background of the task and an overview of our model architecture. Given a point cloud scene $P$ with $N$ points, represented as $P = \{p_0, p_1, \dots, p_{N-1}\}$, where each point $p_i \in \mathbb{R}^{3+d}$ is described by its geometric coordinates $\{x, y, z\}$ and additional features (e.g., RGB color $\{r, g, b\}$, in which case $d=3$). We encode the whole point cloud as the vision input and prompt the LLM to perform different scene understanding tasks on 3D point cloud data. For example, we issue queries like ``Can you segment the \{category\} objects in this point cloud?" to enable open-vocabulary 3D semantic segmentation. 

Figure \ref{fig:main_architecture} presents the overall architecture of the proposed model, which segments point clouds by instructions through three main components: ($\mathit{i}$) Our OcDR, generated by an instance segmentor $\mathcal{E}$ with a cross-attention module $\mathcal{C}$. ($\mathit{ii}$) A Large Language Model $\mathcal{F}$ serves as the core reasoning module. ($\mathit{iii}$) Our GRD that extracts dense interpretations from compact cross-modal features. The OcDR serves as the bridge of the whole pipeline. In the following sections, we illustrate each component of the model, as well as details of the distractor-supervised mechanism of our OcDR.

\subsection{Model Architecture}

\subsubsection{Object-centric Discriminative Representation}

Our OcDR is composed of the class-agnostic object proposals $\mathbf{f}_{q}$ and the point-wise feature $\mathbf{f}_{point}$ for dense information preservation. Both are generated from a pretrained proposal generator. We adopt a cross-attention mechanism $\mathcal{C}$ to aggregate the object information and the dense spatial information, using the $\mathbf{f}_{q}$ as queries and the $\mathbf{f}_{point}$ as keys and values. The proposal generator is kept frozen during training while the cross-attention layer is trainable. In general, we formalize the process above as:
\begin{gather}
    \mathbf{f}_{q}, \mathbf{f}_{point} = \mathcal{E}(P); \
    \mathbf{f}_{OC} = \mathcal{C}(\mathbf{f}_{q}, \mathbf{f}_{point}) .
\end{gather}
The $\mathbf{f}_{q}$ captures target-level semantics and the $\mathbf{f}_{point}$ preserves dense information that will be reactivated through the GRD. The resulting object-point integrated OcDR is then seamlessly connected to the LLM’s input. We generate a fixed set of 150 tokens in all experiments. Our distractor-supervised mechanism is built upon the object-centric structure of our OcDR, and will be introduced in Section \ref{sec:distractors}.

\subsubsection{Multi-Modal Large Language Model}

The input to the Large Language Model (LLM) consists of the $\mathbf{f}_{OC}$, generated by the proposal generator $\mathcal{E}$ (and $\mathcal{C}$), and a text $T$. The text input is first incorporated into a carefully crafted prompt, for instance, \textit{``Can you segment the \{text\} objects in this point cloud?”}, where \textit{\{text\}} serves as a placeholder for the class name or description. This prompt is then tokenized by the LLM tokenizer, $\mathcal{F}_{tokenizer}$, producing the text embeddings $\mathbf{f}_{txt}$ as input for the LLM. This process is formalized as:
\begin{equation}
    \mathbf{f}_{txt} = \mathcal{F}_{tokenizer}(T).
\end{equation}
To effectively integrate multi-modal information into the LLM and strengthen its capability in segmentation tasks, we expand the LLM’s vocabulary by introducing a vision token, \texttt{[point]}, and a special segmentation token, \texttt{[SEG]}, in line with the 2D segmentation paradigm \cite{lai2024lisa, rasheed2024glamm}. With these preparatory modifications, the $\mathbf{f}_{OC}$ and $\mathbf{f}_{txt}$ are then processed by the LLM $\mathcal{F}$ to generate the predicted output. This process can be formalized as follows:
\begin{equation}
    \mathbf{y}_{pred} = \mathcal{F}(\mathbf{f}_{OC}, \mathbf{f}_{txt}).
\end{equation}
During LLM processing, the \texttt{[point]} token will be padded by the $\mathbf{f}_{OC}$. As illustrated in Figure \ref{fig:main_architecture}, after recognizing the \texttt{[SEG]} token, the LLM will output a predicted token for segmentation within the output sequence $\mathbf{y}_{pred}$, donated as $\mathbf{y}_{\texttt{[SEG]}}$. The predicted $\mathbf{y}_{\texttt{[SEG]}}$ is then fed into the next stage for mask generation. We adopt the LLaMA2-7B \cite{touvron2023llama2} as the LLM and apply LoRA \cite{hu2022lora} for efficient fine-tuning.

\subsubsection{Geometric Reactivation Decoder}

With the cross-modal features from the LLM, preserving both target-level semantic and dense features, we design the GRD $\mathcal{D}$ to extract the compact interpretations to produce the required binary mask prediction $\mathbf{m} \in \mathbb{R}^{N}$. As shown in Figure \ref{fig:main_architecture}, our decoder takes a set of learnable mask queries $\mathbf{y}_{\texttt{Mask}}$ as input, and additionally attends to both the $\mathbf{f}_{OC}$ and the LLM predicted output $\mathbf{y}_{\texttt{[SEG]}}$.

The decoder processes mask queries $\mathbf{y}_{\texttt{Mask}}$ through a sequence of attention layers to generate output queries $\hat{\mathbf{y}}_{\texttt{Mask}}$, which are used for object masks. The mask queries first go through a multi-head self-attention mechanism, allowing interaction among queries to refine their representations. The queries are then passed through two multi-head cross-attention layers. The first cross-attention layer attends to the LLM output $\mathbf{y}_{\texttt{[SEG]}}$ (as keys and values), integrating compact cross-modal features and retrieving the object referenced by the language instruction. The second cross-attention layer attends to the OcDR $\mathbf{f}_{OC}$. By revisiting the $\mathbf{f}_{OC}$, the decoder reactivates the preserved details information of the whole scene, identifying those surrounding the target object and reinforcing inter-object relationships. The decoder is constructed by stacking these layers $n$ times, which can be formalized as follows:
\begin{equation}
    \hat{\mathbf{y}}_{\texttt{Mask}} = \mathcal{D}(\mathbf{y}_{\texttt{Mask}}, \mathbf{f}_{OC}, \mathbf{y}_{\texttt{[SEG]}}).
\end{equation}
The output masks are computed by the dot product between the point-wise feature $\mathbf{f}_{point}$ and the output queries $\hat{\mathbf{y}}_{\texttt{Mask}}$. This process is formulated as follows:
\begin{equation}
    \mathbf{m} = \hat{\mathbf{y}}_{\texttt{Mask}} \cdot \mathbf{f}_{point}^\texttt{T}, \mathbf{m} \in \mathbb{R}^{N},
\end{equation}
where the superscript $^\texttt{T}$ represents the matrix transpose.

In addition to the mask output, we also use an MLP-based box head to predict the bounding boxes of the objects. After that, the output query set is further attended to the LLM output $\mathbf{y}_{\texttt{[SEG]}}$ once more, refining their alignment with the target. They then pass through an MLP-based target head, which predicts whether each query contains the target object. By selecting the positive target output masks, we obtain the final segmentation results corresponding to the specified object. Our GRD supports flexible segmentation of one or multiple objects, making it well-suited for real-world applications.

\subsection{Supervision with Distractors \& Matching Loss} 
\label{sec:distractors}
Existing methods fail to distinguish target objects from similar distractors, limiting their effectiveness in dense prediction. Therefore, we introduce a \textbf{distractor-supervised mechanism} that leverages spatial and semantic relationships to refine object differentiation. By explicitly incorporating hard negative distractors—objects with semantic proximity to the target—as supervision signals, the model learns to enhance object identity discrimination, leading to more accurate and robust segmentation.

Specifically, the generated output masks from our GRD will then be matched with the ground truth of both the target objects and our designed distractors. The distractors are defined as objects with semantic proximity to the target, \textit{i.e.}, those objects with the same or akin semantic category as the targets. For instance, when there is more than one \textit{Chair} in the scene, or similar categories, like \textit{Bed} and \textit{Sofa Bed} are present in one scene. We extract these distractors from the semantic categories of ScanNet200 \cite{rozenberszki2022scannet200} and construct a \textbf{semantic distractor dataset} based on open-vocabulary and referring segmentation benchmarks.

\paragraph{Training Loss} With both the target masks and the distractor masks, we match them with the output masks using Hungarian matching \cite{kuhn1955hungarian}. Let us denote the matching objective as $\mathbf{y}$, and the prediction set with $M$ masks as $\hat{\mathbf{y}} = \{\hat{\mathbf{y}_i}\}_{i=1}^{M}$, and the $i$-th prediction is represented by: $~\hat{\mathbf{y}}_i = \left\{ \mathbf{m}_i, \mathbf{c}_i \right\}, $
where $\mathbf{m}_i \in \mathbb{R}^{N}$ is the predicted binary mask and $\mathbf{c}_i \in \mathbb{R}^{2}$ represents the probability scalars indicating the mask categories predicted by the target head (with class 0 indicating the target objects and class 1 indicating the distractors). The Hungarian matching is preceded by finding the best prediction as the positive target/distractor via minimizing the matching cost:
\begin{gather}
    \hat{\mathbf{y}}_{pos} = argmin \ \mathfrak{L}_{match}(\hat{\mathbf{y}}_i, \mathbf{y}); \\
    \mathfrak{L}_{match} = \lambda_{\texttt{cls}}\mathfrak{L}_{\texttt{cls}} + \lambda_{\texttt{mask}}\mathfrak{L}_{\texttt{mask}}.
\end{gather}
Here the $\mathfrak{L}_{\texttt{cls}}(\hat{\mathbf{y}}_i, \mathbf{y})$ is the focal loss \cite{lin2017focal} to supervise the $\mathbf{c}_i$ from the target head, and the $\mathfrak{L}_{\texttt{mask}}(\hat{\mathbf{y}}_i, \mathbf{y})$ sums up the DICE loss \cite{milletari2016dice} and the binary mask focal loss to supervise the output mask $\mathbf{m}_i$. Besides the Hungarian matching, we supervise the predicted bounding box with additional box-related loss $\mathfrak{L}_{\texttt{box}}$ by summing up the L1 loss and the GIoU loss \cite{rezatofighi2019giou}. The predicted bounding box $\mathbf{b}_i \in \mathbb{R}^{6}$ is defined by the center coordinates ($\mathbb{R}^3$) and the width, height, and depth of the box. During training, only the matched $\hat{\mathbf{y}}_{pos}$ contributes to the loss computation. The overall loss is formalized as follows:
\begin{equation}
    \mathfrak{L} = \mathfrak{L}_{\texttt{match}} + \lambda_{\texttt{box}}\mathfrak{L}_{\texttt{box}}.
\end{equation}
Additional details of the loss definition are provided in the \textbf{APPENDIX. A}

\subsection{Training Pipeline}

\paragraph{Prompt Template}
Our model is trained end-to-end with prompts that supervise both open-vocabulary segmentation and referring segmentation, ensuring robust generalization across diverse object categories and language instructions. We design sets of task-specific prompt templates, shown in Table \ref{tab:prompts}. During training, one prompt is randomly sampled from the corresponding prompt set at each iteration to increase linguistic diversity. During evaluation, we use the first one in each list for all methods to ensure stable results.

\begin{table}[htbp]
\caption{\textbf{Prompt template for PLM training.}}
\vspace{-5pt}
\small
\centering
\renewcommand{\arraystretch}{1.3}
{
    \begin{tcolorbox}[colframe=black!30, boxrule=0.5pt, arc=2mm]
    \textbf{Semantic Segmentation:} \\
    \texttt{[point]} Can you segment the \{class\_name\} category in this point cloud? \\ 
    \texttt{[point]} Please segment the \{class\_name\} category in this point cloud. \\
    \texttt{[point]} What is \{class\_name\} category in this point cloud? Please respond with segmentation mask. \\
    \texttt{[point]} What is \{class\_name\} category in this point cloud? Please output segmentation mask. \\
    \textbf{Referring Segmentation:} \\
    \texttt{[point]} With a description: \{description\} Please respond with segmentation mask. \\
    \texttt{[point]} Giving the referring sentence: \{description\} Please output segmentation mask. \\
    \texttt{[point]} Where is the object: \{description\} Can you segment the described object? \\
    \texttt{[point]} In this scene: \{description\} Please segment the described object.
    \end{tcolorbox}
}
\label{tab:prompts}
\vspace{-15pt}
\end{table}

\paragraph{Training Data} We aggregate essential training data from the referring point cloud segmentation benchmark, including ScanRefer \cite{chen2020scanrefer}, ReferIt3D \cite{achlioptas2020referit3d} (both Nr3D and Sr3D), and Multi3DRefer \cite{zhang2023multi3drefer}. We also aggregate ScanNet200 \cite{rozenberszki2022scannet200} from the open vocabulary segmentation benchmark, which also serves as the foundational scene dataset for the aforementioned referring benchmarks. Following previous works on 3D segmentation \cite{he2025segpoint, huang2024openins3d}, we adopt mIoU as the primary metric, defined as the average of all per-point cloud scene Intersection-over-Union (IoU). Additionally, we report AP@25 and AP@50, which are based on IoU thresholds of 25\% and 50\%, respectively, measuring the accuracy of instance predictions at different levels of overlap with ground truth. Our baselines were trained and evaluated under identical data augmentations and prompt settings to ensure a fair comparison. Specifically, we adopt a unified augmentation pipeline for all training, including random point dropout, rotating and flipping, scaling, jittering, and color augmentations. \textbf{Noticed that we did not apply any test-time augmentation in our experiments.}

\paragraph{Implementation Details}
\label{papag:implementation}
We train our models on 4 Nvidia A100 GPUs. The training batch size on each GPU is set to 12, with an initial learning rate of 3e-4. We apply a warm-up for the initial 2\% of training steps, followed by a linear decay of the learning rate. The total training steps are set to 5000 (approximately one epoch). For the model setting, we use the pre-trained class-agnostic Mask3D \cite{schult2023mask3d} to generate 150 object-centric tokens. We adopt the LLaMA2-7B \cite{touvron2023llama2} as the LLM, setting the LORA rank to 8 and the LORA alpha to 16. We initialize 16 mask queries by default to cover most of the targets and distractors. For the GRD, the depth $n$ is set to 3 by default, with the transformer hidden dimension set to 256 and the number of heads set to 8.

\paragraph{Distractor design}

For the distractor setting, we use two straightforward designs to generate the distractor list for each sample. Firstly, for the semantic segmentation task, distractors are constructed by identifying semantically similar categories within the predefined class set (e.g., Bed vs. Sofa Bed, Office chair vs. Armchair). When objects from such semantically overlapping categories occur in the same scene, they are treated as distractors. \textbf{Noted that this process is mutual.} We provide further examples in Table \ref{tab:distractors}. Secondly, for the referring segmentation task, distractors are constructed by identifying the object with the same category as the targets. We provide an example in Figure \ref{fig:example}, among the presented 10 chairs, only one is the target, and the other are distractors.

\begin{table}[bp]
\centering
\caption{Examples for semantic distractors.}
\label{tab:distractors}
\begin{tabular}{l|l}
\toprule
Target categorie & Distractor categories \\
\midrule
shower wall & closet wall \\
office chair & armchair, sofa chair, folded chair \\
tissue box & mailbox \\
\bottomrule
\end{tabular}
\end{table}

\section{Experiments}

To validate the performance of our Point Linguist Model, we conduct extensive experiments on different benchmarks and tasks. We first apply the PLM to the open vocabulary segmentation tasks on ScanNetV2 \cite{dai2017scannet}, ScanNet200 \cite{rozenberszki2022scannet200}, and S3DIS \cite{armeni2016s3dis} datasets. We report results of the open vocabulary instance segmentation (OVIS) and open vocabulary semantic segmentation (OVSS) on those benchmarks to provide a wide range of comparison. Additionally, we perform referring expression segmentation (RES) on the ScanRefer \cite{chen2020scanrefer} and the ReferIt3D \cite{zhang2023multi3drefer} (both Sr3D and Nr3D) datasets. We further evaluate the PLM on the generalized referring expression segmentation (GRES) tasks with the MultiRefer3D \cite{zhang2023multi3drefer} benchmark, which requires a flexible segmentation of zero, one, or multiple objects from the text. We then conduct enriched ablation studies on different components of our model. We also show the model's capability in reasoning and comprehending implicit user instructions. 

\subsection{Compare to State-of-the-Art}

\paragraph{Open-Vocabulary Instance Segmentation} We compare our method with existing 3D open-vocabulary models in the same setting. It is worth noting that the performance on S3DIS is achieved in a zero-shot setting, without scenes seen during training. For a fair comparison, we conduct OVIS while excluding the ``other furniture'' class from ScanNetV2 and the ``clutter'' class from S3DIS, as their definitions are ambiguous. As a result, ScanNetV2 includes 17 classes, and S3DIS contains 12 classes. We follow the category splits from previous methods \cite{ding2023pla, huang2024openins3d} and compare performance on novel classes. As shown in Table \ref{tab:ovis}, our method demonstrates impressive performance in OVIS, surpassing all previous methods in all 3 novel class settings. This proves the effectiveness of language instruction with the LLMs. Specifically, PLM consistently outperforms the prior state-of-the-art method, OpenIns3D \cite{huang2024openins3d}, across all benchmarks.

\begin{table*}
\caption{Open Vocabulary Instance Segmentation results on SacnNetV2 \cite{dai2017scannet} and S3DIS \cite{armeni2016s3dis}. \textbf{B/N} represent \textit{Base} and \textit{Novel} class splitting in the open vocabulary setting. We follow the standard split presented in previous methods \cite{ding2023pla, yang2024regionplc, wang2025xmask3d}.}
\vspace{-5pt}
\begin{center}
{
\begin{tabular}{l|ccc|ccc}
\toprule
    \multirow{2}{*}{Methods} & \multicolumn{3}{c|}{ScanNetV2} & \multicolumn{3}{c}{S3DIS} \\
    & B/N & AP50 & AP25 & B/N & AP50 & AP25 \\
\midrule
    PLA \cite{ding2023pla} & 10 / 7 & 21.9 &  - & 8 / 4 & 8.6 & - \\
    RegionPLC \cite{yang2024regionplc} & 10 / 7 &  32.3 & - & 8 / 4 & - & - \\
    Lowis3D \cite{ding2024lowis3d} & 10 / 7 &  31.2 & - & 8 / 4 & 13.8 & - \\
    Open3DIS \cite{nguyen2024open3dis} & 10 / 7 & - & -	& 8 / 4 & 26.3 & - \\
    Mask3d+PointClip \cite{zhu2023pointclip2} & - / 7 &  4.5 &  7.8 & - / 4 & 5.4 & 10.3 \\
    OpenIns3D \cite{huang2024openins3d} & - / 7 & 27.9 & 42.6 & - / 4 & 37.0 & 39.3 \\
    \textbf{PLM} & - / 7 & \textbf{54.1} & \textbf{62.4} & - / 4 &  \textbf{38.7} &  \textbf{52.6} \\
\midrule
    PLA \cite{ding2023pla} & 8 / 9 & 25.1 &  - & 6 / 6 & 9.8 & - \\
    RegionPLC \cite{yang2024regionplc} & 8 / 9 &  32.2 & - & 6 / 6 & - & - \\
    Lowis3D \cite{ding2024lowis3d} & 8 / 9 &  38.1 & - & 6 / 6 & 15.8 & -\\
    Open3DIS \cite{nguyen2024open3dis} & 8 / 9 & - & -	& 6 / 6 & 29.0 & - \\
    Mask3d+PointClip \cite{zhu2023pointclip2} & - / 9 &  5.6 &  6.7 & - / 6 &  8.5 & 10.6 \\
    OpenIns3D \cite{huang2024openins3d} & - / 9 & 19.5 & 27.9 & - / 6 & 33.0 & 38.9\\
    \textbf{PLM} & - / 9 & \textbf{60.5} & \textbf{73.2} & - / 6 & \textbf{34.0} &  \textbf{43.5} \\
\midrule
    Mask3d+PointClip \cite{zhu2023pointclip2} & - / 17 &  4.5 &  14.4 & - / 12 &  8.6 & 9.3 \\
    OpenIns3D \cite{huang2024openins3d} & - / 17 & 28.7 & 38.9 & - / 12 & 28.3 & 29.5\\
    \textbf{PLM} & - / 17 & \textbf{38.4} & \textbf{46.2} & - / 12 & \textbf{29.3} & \textbf{35.4} \\
\bottomrule
\end{tabular}
}
\end{center}
\label{tab:ovis}
\vspace{-15pt}
\end{table*}

\begin{table}[htbp]
\caption{Open vocabulary semantic segmentation result in ScanNetV2 \cite{dai2017scannet} and ScanNet200 \cite{rozenberszki2022scannet200}. We report the mIoU result.}
\vspace{-10pt}
\begin{center}
{
\begin{tabular}{l|c|c}
\toprule
    Methods & ScanNet & ScanNet200 \\
\midrule
    ConceptFusion \cite{jatavallabhula2023conceptfusion} & 33.3 & 8.8 \\
    OpenMask3D \cite{takmaz2023openmask3d} & 34.0 & 10.5 \\ 
    OpenScene-2D \cite{peng2023openscene} & 41.4 & 12.7 \\
    OpenScene-3D \cite{peng2023openscene} & 46.0 & 6.3 \\
    OpenScene-2D/3D \cite{peng2023openscene} & 47.5 &  11.6  \\
    Diff2Scene \cite{zhu2024opendiffusion} & 48.6 & 14.2 \\
\midrule
    \textbf{PLM}  & \textbf{66.0} & \textbf{43.5}  \\
\bottomrule
\end{tabular}
}
\end{center}
\label{tab:ovss}
\vspace{-10pt}
\end{table}

\begin{table}[htbp]
\caption{Open vocabulary semantic segmentation on different novel class settings. We report the mIoU result of the standard split following previous methods.}
\vspace{-5pt}
\begin{center}
{
\begin{tabular}{l|ccc|cc|cc}
\toprule
    \multirow{2}{*}{Methods} & \multicolumn{3}{c|}{ScanNetV2} & \multicolumn{2}{c|}{ScanNet200} & \multicolumn{2}{c}{S3DIS} \\
    & N4 & N7 & N9 & N30 & N50 & N4 & N6 \\
\midrule
    LSeg-3D \cite{li2022lseg3d}  & 0.0 & 0.1 & 0.9 & 0.8 & 1.6 & 0.1 & 0.0 \\
    3DGenZ \cite{michele20213dgz}  & 12.6 & 13.3 & 6.6 & 1.4 & 1.9 & 4.8 & 6.1 \\
    3DTZSL \cite{cheraghian20203dtzsl}  & 6.1 & 2.0 & 4.2 & 0.5 & 0.4 & 4.7 & 1.9 \\
    PLA \cite{ding2023pla} & 62.4 & 45.9 & 40.8 & 7.8 & 6.6 & 24.5 & 29.4 \\
    OpenScene \cite{peng2023openscene} & 62.8 & 51.7 & 43.6 & 10.4 & 11.2 & 33.2 & 36.4 \\
    XMask3D \cite{wang2025xmask3d} & 70.2 & 55.1 & 43.8 & 13.3 & 11.4 & 37.2 & 39.1 \\
\midrule
    \textbf{PLM}  & 70.0 & \textbf{66.1} & \textbf{66.5} & \textbf{42.5} & \textbf{43.9} & \textbf{51.2} & \textbf{57.5} \\
\bottomrule
\end{tabular}
}
\end{center}
\label{tab:ovssnovel}
\vspace{-15pt}
\end{table}

On ScanNetV2, PLM achieves the highest AP50 and AP25 scores in all settings. Even in the most challenging 17-class setting, PLM shows a strong lead with 38.4\% AP50 and 46.2\% AP25, outperforming OpenIns3D by \textbf{$\uparrow$9.7\%} and \textbf{$\uparrow$7.3\%}, respectively. In a zero-shot way, similar trends are observed in the S3DIS dataset, where PLM consistently achieves the highest performance in all settings. In the 12-class setting, PLM secures the top position with 29.3\% AP50 and 35.4\% AP25, showing a \textbf{1.0\%} and \textbf{5.9\%} improvement over OpenIns3D. We also provide the per-class results in Section \ref{sec:perclass}

\begin{figure*}[htbp]
    \centering
    \includegraphics[width=0.9\linewidth]{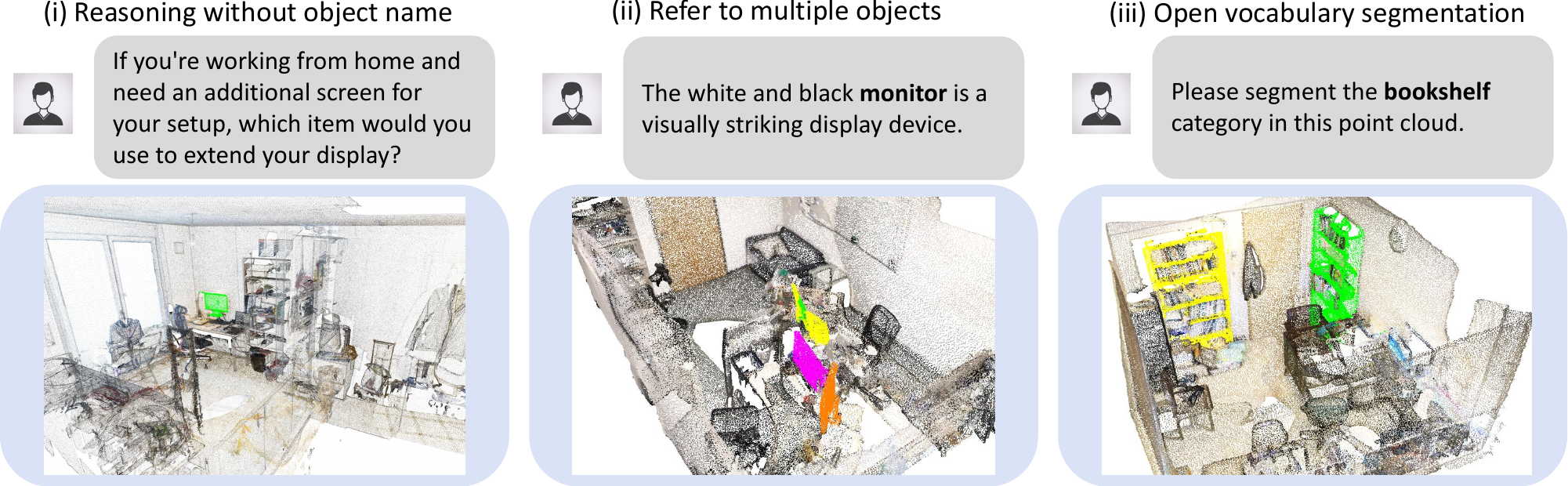}
    \caption{Visualization results of PLM in different segmentation tasks. (\textit{i}) Our model can easily reason and comprehend implicit user instructions. (\textit{ii}) and (\textit{iii}) Our model enables flexible segmentation of multiple objects with clear instance separation. Different highlight colors represent different instances.}
    \label{fig:taskvis}
\vspace{-10pt}
\end{figure*}

\paragraph{Open-Vocabulary Semantic Segmentation} 

To further validate the effectiveness of our proposed PLM, we evaluate its performance on the OVSS task, following the standard evaluation protocols from prior works \cite{wang2025xmask3d, peng2023openscene, zhu2024opendiffusion}. We compare our PLM with the latest state-of-the-art method Diff2Scene \cite{zhu2024opendiffusion} and XMask3D \cite{wang2025xmask3d}. Table \ref{tab:ovss} presents the mIoU results on ScanNetV2 \cite{dai2017scannet} and ScanNet200 \cite{rozenberszki2022scannet200} datasets. Compared to previous methods, PLM demonstrates superior performance across both datasets, achieving 66.0\% mIoU on ScanNetV2 and 43.5\% mIoU on ScanNet200, significantly surpassing previous state-of-the-art approaches. Table \ref{tab:ovssnovel} further breaks down the OVSS results across various novel class splits. We observe that PLM consistently achieves the highest performance across all settings, demonstrating its strong generalization ability in recognizing novel classes in open-vocabulary settings.

\begin{table}
\caption{Referring segmentation result. We conduct the generalized referring expression segmentation on the MultiRefer3D \cite{zhang2023multi3drefer} dataset. And conduct the single-object referring expression segmentation on the ScanRefer \cite{chen2020scanrefer} and ReferIt3D \cite{zhang2023multi3drefer} datasets.}
\vspace{-5pt}
\begin{center}
% \resizebox{\linewidth}{!}
{
\begin{tabular}{l|c|c|c|c}
\toprule
    Methods & Multi3DRefer & ScanRefer & Nr3D & Sr3D \\
\midrule
    M3DRef-CLIP \cite{zhang2023multi3drefer} & 32.6 & 35.7 & 27.0 & - \\
    3D-STMN \cite{wu2024dstmn} & - & 39.5 & 27.6 & 34.4 \\ 
    TGNN \cite{huang2021textguided} & - & 27.8 & 19.1 & 20.2 \\
    BUTD-DETR \cite{jain2022bottom} & 26.2 & 35.4 & 27.5 & - \\
    EDA \cite{wu2023eda} & 28.9 & 36.2 & 29.3 & - \\
    X-RefSeg3D \cite{qian2024xrefsegd} & - & 29.9 & - & -\\
    RefMask3D \cite{he2024refmask3d} & - & \textbf{44.8} & - & - \\
    SegPoint \cite{he2025segpoint} & 36.1 & 41.7 & 32.2 & - \\
\midrule
    \textbf{PLM} & \textbf{42.1} & 43.1 & \textbf{32.4} & \textbf{34.9} \\
\bottomrule
\end{tabular}
}
\end{center}
\label{tab:refer}
\vspace{-10pt}
\end{table}

\paragraph{Referring Expression Segmentation} Table~\ref{tab:refer} compares the results on RES and GRES. Previous LLM-based methods like SegPoint~\cite{he2025segpoint} only output a single segmentation mask. It manages the GRES by combining all target masks as one, resulting in suboptimal performance due to the lack of instance-level differentiation. In contrast, our PLM enables flexible segmentation of multiple objects, achieving state-of-the-art results (\textbf{42.1\%} mIoU) on MultiRefer3D and significantly surpassing SegPoint by \textbf{$\uparrow$6.0\%} and other previous methods, demonstrating its superior ability to understand multi-object references. As for the single-object Referring Expression Segmentation (RES), PLM also demonstrates strong performance across benchmarks. It achieves 43.1\% mIoU on ScanRefer, slightly below the previous expert model RefMask3D (44.8\%), while still maintaining the lead on Nr3D (32.4\%) and Sr3D (34.9\%), surpassing all previous methods.

These results highlight the advantages of our language-guided multi-mask generation, allowing for better instance separation and higher accuracy in referring segmentation tasks. Unlike SegPoint, which merges all target masks into a single prediction, PLM effectively models individual instances, ensuring precise multi-object grounding. This capability is crucial for tasks like OVIS, where PLM leverages LLM-based modeling to handle diverse and novel object categories. In contrast, SegPoint lacks this ability, making it incapable of performing OVIS effectively. This further demonstrates the importance of language-driven feature alignment and structured multi-instance reasoning in 3D open-vocabulary segmentation.

\paragraph{Reasoning Expression Segmentation} As shown in Figure \ref{fig:taskvis} (\textit{i}), PLM is also capable of reasoning and comprehending implicit user instruction, just like LISA \cite{lai2024lisa}. However, our study is constrained by the lack of a standardized benchmark for reasoning expression segmentation. While some works \cite{he2025segpoint, huang2024reason3d, jiang2024reasonseg3d} have introduced related datasets, they remain closed-source as of this writing. Therefore, we qualitatively demonstrate the reasoning capabilities of PLM, with additional samples provided in the \textbf{APPENDIX. B}.

\subsection{Ablation Study}

After extensively demonstrating our superiority across various tasks, we conduct comprehensive ablation studies to validate the rationality of our design choices for the OcDR and GRD. We select the challenging MultiRefer3D and ScanRefer benchmarks as our ablation baselines. 

\paragraph{Ablation on OcDR tokens}

Our OcDR encodes the entire scene in an object-centric manner at the feature level, emphasizing high-level semantic structure rather than precise mask boundaries. The subsequent GRD module refines segmentation masks only at the final stage. As a result, even when some proposals are inaccurate or missing, the overall scene representation remains informative, mitigating sensitivity to proposal errors produced by Mask3D. We analyze the robustness of PLM to the number of OcDR tokens. As shown in Table \ref{tab:ablationoc}, although the encoder is pretrained with 150 tokens, varying the token number to 50 or 200 results in small performance changes. Specifically, using fewer tokens may omit some valid object proposals and lead to a less complete scene representation, while using more tokens may introduce noisier proposals. This indicates that OcDR does not rely on a narrowly tuned token budget and that moderate mismatch between pretraining and inference token counts does not critically affect representation quality. The model maintains stable performance, demonstrating robustness to variations in proposal quality.

\begin{table}[tbp]
\caption{Compare OcDR with other point cloud encoding methods. \colorbox{cyan!25!white}{Blue line} represents the default settings of our PLM.}
\vspace{-10pt}
\begin{center}
% \resizebox{\linewidth}{!}
{
\begin{tabular}{l|c|c}
\toprule
    Representations & Multi3DRefer & ScanRefer \\
\midrule
    OcDR (50 tokens) & 41.2 & 42.0 \\
    \rowcolor{cyan!20} {OcDR (150 tokens)} & 42.1 & 43.1 \\
    OcDR (200 tokens) & 41.3 & 42.2 \\
\midrule
    Uni3D \cite{zhou2023uni3d} (512 tokens) & 32.7 & 35.4 \\
    Uni3D \cite{zhou2023uni3d} (1024 tokens) & 32.9 & 36.3 \\
    mask proposal + Uni3D \cite{zhou2023uni3d} & 35.6 & 37.8 \\
\midrule
    w/o distractor & 38.9 & 40.1 \\
    random distractor & 10.1 & 12.3 \\
\bottomrule
\end{tabular}
}
\end{center}
\label{tab:ablationoc}
\vspace{-10pt}
\end{table}

\begin{table}[tbp]
\centering
\caption{Comparison of data efficiency with 50\% of training data.}
% \vspace{-5pt}
\label{tab:lessdata}
% \resizebox{\linewidth}{!}
{
\begin{tabular}{l|c|c}
\toprule
Methods & Multi3DRefer & ScanRefer \\
\midrule
SegPoint \cite{he2025segpoint} & 36.1 & 41.7 \\
\textbf{PLM} (50\% data) & 40.9 & 41.2 \\
\bottomrule
\end{tabular}
}
\vspace{-5pt}
\end{table}

\paragraph{Ablation on OcDR Design}

We compare our OcDR with two existing point cloud encoding methods used in the SegPoint \cite{li2022lseg3d} and Chat-Scene \cite{huang2023chatscene}.

Firstly, SegPoint \cite{li2022lseg3d} relies on dense point patches generated by the pretrained Uni3D backbone \cite{zhou2023uni3d}. Comparing the performance on the downstream tasks, results in Table \ref{tab:ablationoc} reveal that using the SegPoint method with Uni3D directly leads to a significant performance drop (-6.5\% on MultiRefer3D, -5.3\% on ScanRefer), confirming that our structured OcDR provides better object grounding and segmentation quality. While increasing the token length from 512 to 1024 slightly improves ScanRefer performance (+0.9\%), the results remain far below our setting, highlighting the limitations of relying solely on excessive tokenization without structured OcDR. 

Secondly, Chat-Scene \cite{huang2023chatscene} uses isolated object tokens generated by applying Uni3D \cite{zhou2023uni3d} on the mask proposal from Mask3D \cite{schult2023mask3d}. Similarly, the second setting (mask proposal + Uni3D, in Table \ref{tab:ablationoc}) also suffers from degraded performance. While leveraging Uni3D on mask proposals provides a structured object-centric representation, it lacks holistic scene awareness, leading to a notable performance drop. 

\paragraph{Ablation on Distractor Supervision}

We also do ablation on the distractor-supervised mechanism. We can observe that in the setting without the semantic distractor, the performance notably declines. This demonstrates the importance of semantic distractors in refining the segmentation by enhancing object discrimination and reducing ambiguity in multi-object scenarios. Notably, the semantic similarity design is essential, as replacing the distractor with random objects leads to a drastic performance drop, effectively causing learning to collapse. 

\begin{table}[tbp]
\centering
\caption{Comparison of computation cost on different representations. \colorbox{cyan!25!white}{Blue line} represents the default settings of our PLM.}
\label{tab:computation}
\resizebox{\linewidth}{!}
{
\begin{tabular}{l|c|c}
\toprule
Representations & inferencing time (s) & acceptable batch size \\
\midrule
\rowcolor{cyan!20} OcDR (150 tokens) & 0.28 & 12 \\
Uni3D \cite{zhou2023uni3d} (512 tokens) & 0.51 & 12 \\
Uni3D \cite{zhou2023uni3d} (1024 tokens) & 1.07 & 4 \\
\bottomrule
\end{tabular}
}
\vspace{-5pt}
\end{table}

\begin{table}[tbp]
\caption{Ablation study on different parameter settings for the GRD. \colorbox{cyan!25!white}{Blue line} represents the default settings: Decoder $n$=3; mask queries = 16.}
\vspace{-5pt}
\begin{center}
% \resizebox{\linewidth}{!}
{
\begin{tabular}{c|c|c}
\toprule
     GRD & Multi3DRefer & ScanRefer \\
\midrule
    \rowcolor{cyan!20} default settings & 42.1 & 43.1 \\
\midrule
    w/o Decoder & 27.9 & 32.5 \\
\midrule 
    Decoder $n$=1 & 40.5 & 39.8 \\
    Decoder $n$=5 & 42.1 & 42.8 \\
\midrule
    mask queries = 4 & 41.1 & 40.7 \\
    mask queries = 24 & 41.7 & 42.8 \\
    mask queries = 48 & 41.0 & 38.6 \\
\midrule
    w/o bbox loss & 40.8 & 42.4 \\
\bottomrule
\end{tabular}
}
\end{center}
\label{tab:ablation}
\vspace{-15pt}
\end{table}

\paragraph{Ablation on Data Efficiency}

Our OcDR also demonstrates superior data efficiency. We train our PLM using just 50\% of the training data and benchmark it against Segpoint \cite{he2025segpoint}, the current SOTA point-patch token–based LLM model. Results are shown in Table \ref{tab:lessdata}. We match Segpoint’s performance using only half the training data, and even surpass it on the Multi3DRefer \cite{zhang2023multi3drefer} multi-object segmentation task. Furthermore, we evaluate the model's efficiency, comparing the OcDR with the point-patch tokens. We use excessive point-patch tokens generated by Uni3D \cite{zhou2023uni3d}, a strong point cloud encoder pre-trained with language alignment. Uni3D provides a rich semantic understanding of 3D objects and enables better alignment with LLMs. It serves as the point cloud encoder in the SegPoint \cite{he2025segpoint} model. As shown in Table \ref{tab:computation}, Uni3D incurs a significant computational cost, leading to slower inference and reduced batch efficiency. Increasing token length from 512 to 1024 further exacerbates the issue, more than doubling the inference time while drastically reducing the acceptable batch size. These two results highlight the inefficiency of relying on excessive point-patch tokens, making it impractical for scalable deployment. 

\paragraph{Ablation on GRD Design}
We then evaluate the impact of different settings of the proposed GRD and the distractor-supervised mechanism. Firstly, we test the performance in a setting without the decoder. In this case, we directly use the vanilla output of LLMs as the mask query, generating the final mask by computing the dot product between the point-wise features and the $\mathbf{y}_{\texttt{[SEG]}}$. This process is formulated as follows:
\begin{equation}
    \mathbf{m} = \mathbf{y}_{\texttt{[SEG]}} \cdot \mathbf{f}_{point}^\texttt{T}, \mathbf{m} \in \mathbb{R}^{N}.
\end{equation}
As shown in Table \ref{tab:ablation}, our decoder plays a crucial role in segmentation accuracy, significantly enhancing multi-object segmentation and instance separation.

We also analyze the GRD design choices, including the decoder depth, mask query number, and bounding box loss, and select the default configuration based on consistently robust performance (Table \ref{tab:ablation}). The mask query number implicitly limits the maximum number of distractors participating in matching; when this limit is exceeded, distractors are randomly subsampled. As shown in Table \ref{tab:ablation}, PLM remains stable across different mask query settings, supporting the robustness of the distractor-supervised matching and the GRD design.

\begin{table*}[htbp]
\caption{Per-class results of OVIS on ScanNetV2.  Performance on novel classes is \colorbox{teal!25!white}{highlighted}. The AP50 result is reported.}
% \vspace{-5pt}
\begin{center}
\resizebox{\linewidth}{!}
{
\begin{tabular}{l|c|ccccccccccccccccc}
\toprule
    Methods & B / N
    & \rotatebox{90}{cabinet}
    & \rotatebox{90}{bed} 
    & \rotatebox{90}{chair} 
    & \rotatebox{90}{sofa} 
    & \rotatebox{90}{table} 
    & \rotatebox{90}{door} 
    & \rotatebox{90}{window} 
    & \rotatebox{90}{bookshelf} 
    & \rotatebox{90}{picture} 
    & \rotatebox{90}{counter} 
    & \rotatebox{90}{desk} 
    & \rotatebox{90}{curtain} 
    & \rotatebox{90}{fridge} 
    & \rotatebox{90}{shower c.} 
    & \rotatebox{90}{toilet} 
    & \rotatebox{90}{sink}
    & \rotatebox{90}{bathtub} \\
\midrule 
    \multirow{3}{*}{PLA \cite{ding2023pla}} & 13 / 4 & 50.5 & 77.0 & 82.9 & \cellcolor{teal!15}{43.4} & 75.4 
    & 49.0 & 46.0 & \cellcolor{teal!15}{43.7} & 46.5 & 33.7 & \cellcolor{teal!15}{\textbf{23.2}} & 54.1 & 49.6 & 56.0 & \cellcolor{teal!15}{\textbf{97.8}} & 47.5 & 85.8 \\
    & 10 / 7 & 53.7 & \cellcolor{teal!15}{\textbf{62.7}} & \cellcolor{teal!15}{11.2} & 70.5 & \cellcolor{teal!15}{27.2} & 47.7 & 45.7 & \cellcolor{teal!15}{30.0} & \cellcolor{teal!15}{01.5} & 39.9 & 40.8 & 50.6 & 68.6 & 84.6 & 92.9 & \cellcolor{teal!15}{24.6} & \cellcolor{teal!15}{0.0}   \\
    & 8 / 9 & 45.1 & 77.4 & 82.2 & 84.2 & 74.2 & 48.9 & 51.0 & \cellcolor{teal!15}{30.0} & \cellcolor{teal!15}{0.5} & \cellcolor{teal!15}{2.1} & \cellcolor{teal!15}{16.8} & 44.9 & \cellcolor{teal!15}{\textbf{28.3}} & \cellcolor{teal!15}{35.1} & \cellcolor{teal!15}{94.3} & \cellcolor{teal!15}{16.6} & \cellcolor{teal!15}{0.0}   \\
\midrule
    OpenIns3D \cite{huang2024openins3d} & - / 17 & \cellcolor{teal!15}{24.3} & \cellcolor{teal!15}{52.5} & \cellcolor{teal!15}{\textbf{75.7}} & \cellcolor{teal!15}{\textbf{61.6}} & \cellcolor{teal!15}{40.6} & \cellcolor{teal!15}{39.7} & \cellcolor{teal!15}{45.5} & \cellcolor{teal!15}{\textbf{54.8}} & \cellcolor{teal!15}{0.5} & \cellcolor{teal!15}{\textbf{33.5}} & \cellcolor{teal!15}{16.7} & \cellcolor{teal!15}{\textbf{48.1}} & \cellcolor{teal!15}{18.5} & \cellcolor{teal!15}{4.3} & \cellcolor{teal!15}{50.1} & \cellcolor{teal!15}{16.8} & \cellcolor{teal!15}{7.6}\\
\midrule
    PLM  & - / 17 & \cellcolor{teal!15}{\textbf{30.2}} & \cellcolor{teal!15}{58.6} & \cellcolor{teal!15}{57.5} & \cellcolor{teal!15}{28.7} & \cellcolor{teal!15}{\textbf{41.2}} & \cellcolor{teal!15}{\textbf{45.5}} & \cellcolor{teal!15}{\textbf{47.6}} & \cellcolor{teal!15}{38.8} & \cellcolor{teal!15}{41.2} & \cellcolor{teal!15}{20.4} & \cellcolor{teal!15}{16.9} & \cellcolor{teal!15}{25.4} & \cellcolor{teal!15}{23.0} & \cellcolor{teal!15}{\textbf{36.2}} & \cellcolor{teal!15}{22.0} & \cellcolor{teal!15}{\textbf{55.8}} & \cellcolor{teal!15}{\textbf{64.3}}  \\
\bottomrule
\end{tabular}
}
\end{center}
\label{tab:perclass-scannet}
% \vspace{-10pt}
\end{table*}

\begin{table*}[htbp]
\caption{Per-class results of OVIS on S3DIS. Performance on novel classes is \colorbox{teal!25!white}{highlighted}. The AP50 result is reported.}
% \vspace{-5pt}
\begin{center}
% \resizebox{\linewidth}{!}
{
\begin{tabular}{l|c|cccccccccccc}
\toprule
    Methods & B / N
    & \rotatebox{90}{ceiling}
    & \rotatebox{90}{floor} 
    & \rotatebox{90}{wall} 
    & \rotatebox{90}{beam} 
    & \rotatebox{90}{column} 
    & \rotatebox{90}{window} 
    & \rotatebox{90}{door} 
    & \rotatebox{90}{table} 
    & \rotatebox{90}{chair} 
    & \rotatebox{90}{sofa} 
    & \rotatebox{90}{bookcase} 
    & \rotatebox{90}{board} \\
\midrule 
    \multirow{3}{*}{PLA \cite{ding2023pla}} & 8 / 4 & 89.5 & 100.0 & 50.8 & 0.0 & 35.3 & \cellcolor{teal!15}{36.2} & 60.5 & \cellcolor{teal!15}{0.1} & 84.6 & \cellcolor{teal!15}{1.9} & \cellcolor{teal!15}{0.8} & 59.4 \\
    & 6 / 6 & 89.5 & \cellcolor{teal!15}{60.2} & 17.9 & 0.0 & 41.5 & \cellcolor{teal!15}{10.2} & \cellcolor{teal!15}{2.1} & \cellcolor{teal!15}{0.6} & 86.2 & 45.1 & \cellcolor{teal!15}{0.1} & \cellcolor{teal!15}{2.2} \\
\midrule
    OpenIns3D \cite{huang2024openins3d} & - / 12 & \cellcolor{teal!15}{0.0} & \cellcolor{teal!15}{\textbf{84.4}} & \cellcolor{teal!15}{\textbf{29.0}} & \cellcolor{teal!15}{0.0} & \cellcolor{teal!15}{0.0} & \cellcolor{teal!15}{\textbf{62.6}} & \cellcolor{teal!15}{25.2} & \cellcolor{teal!15}{25.5} & \cellcolor{teal!15}{52.0} & \cellcolor{teal!15}{\textbf{60.0}} & \cellcolor{teal!15}{0.0} & \cellcolor{teal!15}{0.0} \\
\midrule
    PLM  & - / 12 & \cellcolor{teal!15}{0.0} & \cellcolor{teal!15}{0.0} & \cellcolor{teal!15}{0.8} & \cellcolor{teal!15}{0.0} & \cellcolor{teal!15}{0.0} & \cellcolor{teal!15}{40.7} & \cellcolor{teal!15}{\textbf{52.8}} & \cellcolor{teal!15}{\textbf{25.6}} & \cellcolor{teal!15}{\textbf{67.6}} & \cellcolor{teal!15}{51.6} & \cellcolor{teal!15}{\textbf{25.9}} & \cellcolor{teal!15}{\textbf{57.0}} \\
\bottomrule
\end{tabular}
}
\end{center}
\label{tab:perclass-s3dis}
% \vspace{-10pt}
\end{table*}

\begin{table*}[htbp]
\centering
\caption{Computational performance of our PLM.}
\label{tab:grdcomputation}
\begin{tabular}{lccccccc}
\toprule
\multirow{2}{*}{Setting} & \multicolumn{3}{c}{\textbf{Computation (TFLOPs)}} & \multicolumn{2}{c}{\textbf{Throughput}} &
\multirow{2}{*}{\textbf{Memory (GB)}} \\
\cmidrule(lr){2-4} \cmidrule(lr){5-6}
& forward & backward & total & samples/sec & latency (ms) \\
\midrule
Decoder $n$=3 & 47.19 & 94.37 & 141.56 & 3.2 & 312.76 & 17.4 \\
Decoder $n$=5 & 48.79 & 97.57 & 146.36 & 3.1 & 326.79 & 17.4 \\
\bottomrule
\end{tabular}
\end{table*}

\paragraph{Ablation on Hungarian Matching}

We then do ablation on the Hungarian loss weights. The Hungarian matching involves five loss terms, including the classification loss, mask focal loss, Dice loss, L1 loss, and GIoU loss. We adopt the empirical weight setting from Mask3D without specific fine-tuning, and additionally evaluate a uniform-weight variant. The results are reported in Table \ref{tab:Hungarian}.

\begin{table}[htbp]
\centering
\caption{Ablation on Hungarian loss weight settings. The loss weights are reported in the order of classification loss, mask focal loss, Dice loss, L1 loss, and GIoU loss.}
\vspace{-5pt}
\begin{center}
\begin{tabular}{l|c|c}
\toprule
    loss weights & Multi3DRefer & ScanRefer \\
\midrule
    2,2,5,1,2 & 42.1 & 43.1 \\
    1,1,1,1,1 & 39.4 & 40.0 \\
\bottomrule
\end{tabular}
\end{center}
\label{tab:Hungarian}
\vspace{-10pt}
\end{table}

As expected, the loss weights have a noticeable impact on performance, since the accuracy of Hungarian matching directly affects how well the target object is distinguished from distractors. This behavior is inherent to matching-based frameworks rather than specific to our method. Importantly, our key contribution lies not in tuning these weights, but in introducing distractor-aware matching to explicitly differentiate target objects from semantically similar distractors.

\subsection{Per-class results} 
\label{sec:perclass}
We present per-class results on the OVIS task in Table \ref{tab:perclass-scannet} and Table \ref{tab:perclass-s3dis}. The results are reported in terms of AP50, with performance on novel classes highlighted in teal. On the ScanNetV2 dataset, PLM consistently outperforms OpenIns3D across most novel classes, demonstrating its superior generalization ability in the open-vocabulary setting. Compared to PLA \cite{ding2023pla}, PLM achieves more stable and higher recognition rates for underrepresented categories, especially in challenging categories like picture (41.2\%), counter (20.4\%), and desk (16.9\%). PLM outperforms OpenIns3D in key furniture categories, such as cabinet (30.2\% vs. 24.3\%) and bathtub (64.3\% vs. 7.6\%), indicating its stronger ability to segment complex object classes. On the S3DIS dataset, PLM achieves competitive performance in architectural components and furniture classes, such as window (40.7\%), door (52.8\%), table (25.6\%), and board (57.0\%). Compared to OpenIns3D, PLM improves segmentation performance on chairs (67.6\% vs. 52.0\%) and bookcases (25.9\% vs. 0.0\%), demonstrating strong semantic transferability in the open-vocabulary setting.

\subsection{Computational Performance of the PLM}

For modern multimodal LLM-based methods, the LLM component typically dominates the overall computational cost. Since different approaches may adopt different LLM backbones, computational comparisons across methods are often less indicative of the efficiency of the proposed 3D reasoning modules themselves.

Under these constraints, we aim to provide meaningful efficiency analysis from multiple complementary perspectives. First, to offer a coarse latency reference, we compare our inference latency with representative non-LLM-based methods, including OneFormer3D \cite{Kolodiazhnyi2024Oneformer} (221.00 ms), UniSeg3D \cite{xu2024unified} (230.03 ms), SSTNet \cite{liang2021instance} (400 ms), and PointGroup \cite{jiang2020pointgroup} (372 ms). While these methods are not directly comparable due to architectural differences, the comparison provides a useful sense of the practical runtime scale.

We then report a detailed computational analysis of our model in Table \ref{tab:grdcomputation}. Specifically, we report forward, backward, and total computation (in TFLOPs), throughput in samples per second, per-sample latency, as well as GPU memory usage. All results are measured using a batch size of 12 under the same hardware configuration. Under the default setting with GRD Decoder depth $n$=3, our model achieves a throughput of 3.2 samples/sec with a latency of 312.76 ms. To further examine the computational efficiency of our design, we analyze the impact of increasing the GRD depth. As shown in Table \ref{tab:grdcomputation}, increasing the GRD depth from $n$=3 to $n$=5 results in only a marginal increase in total computation (141.56 → 146.36 TFLOPs), while the throughput (3.2 → 3.1 samples/sec) and latency (312.76 → 326.79 ms) remain largely unchanged.

\section{Conclusion}

This paper presented the Point Linguist Model, which bridges the gap between 3D LLMs and dense object segmentation by leveraging novel object-centric representations. Through a meticulously designed GRD and a distractor-supervised mechanism, our model achieved state-of-the-art performance across 7 benchmarks spanning 4 diverse point cloud segmentation tasks. Beyond advancing the state-of-the-art of point cloud segmentation learning, this work paves the way for future research into efficient and effective representations for object-oriented 3D MLLMs. Our future work aims to break the limitations of point cloud segmentation by enhancing object-centric representations within multi-modal LLMs (MLLMs) for advanced 3D reasoning. We further seek to broaden the scope of 3D scene understanding, enabling more generalized and adaptable 3D perception systems.

\appendix

\subsection{Additional Loss Details}

\label{sec:addloss}
\paragraph{Notation}
\begin{itemize}
  \item $M$: Number of predicted masks.
  \item $N$: Number of points per mask.
  \item $\mathbf{y} = \{\mathbf{y}_j\}_{j=1}^M$: Set of ground-truth masks (including both targets and distractors), where each $\mathbf{y}_j$ comprises a binary mask and a class label.
  \item $\hat{\mathbf{y}} = \{\hat{\mathbf{y}}_i\}_{i=1}^M$: Set of predictions, where
    \[
      \hat{\mathbf{y}}_i = \bigl\{\mathbf{m}_i,\;\mathbf{c}_i,\;\mathbf{b}_i\bigr\},
    \]
    with
    \begin{itemize}
      \item $\mathbf{m}_i = (p_i^1, \dots, p_i^N)\in[0,1]^N$: predicted mask probability vector,
      \item $\mathbf{c}_i = [p_{i,0},\,p_{i,1}]^\top\in[0,1]^2$: predicted class probabilities ($0$ = target, $1$ = distractor),
      \item $\mathbf{b}_i\in\mathbb{R}^6$: predicted bounding box, parameterized by center $(x,y,z)\in\mathbb{R}^3$ and size $(w,h,d)\in\mathbb{R}^3$.
    \end{itemize}
  \item $\mathbf{y}_{\text{mask}}=(g^1,\dots,g^N)\in\{0,1\}^N$: ground-truth mask vector.
  \item $\mathbf{y}_{\text{cls}}=[y_0,y_1]^\top\in\{0,1\}^2$: one-hot class label.
  \item $B_i$ and $B_{\text{gt}}$: volumes of the predicted and ground-truth boxes, respectively; $C$ is the smallest enclosing box of $B_i\cup B_{\text{gt}}$, with volume $|C|$.
  \item Hyperparameters $\lambda_{\text{cls}},\;\lambda_{\text{mask}},\;\lambda_{\text{box}}$: weights for the classification, mask, and box losses.
\end{itemize}

\paragraph{Hungarian Matching}
We perform one-to-one matching between $\hat{\mathbf{y}}$ and $\mathbf{y}$ via the Hungarian algorithm:
\[
\begin{split}
  \hat{\mathbf{y}}_{\text{pos}}
  &= \arg\min\;\mathfrak{L}_{\text{match}}\bigl(\hat{\mathbf{y}}_i,\mathbf{y}\bigr), \\
  \mathfrak{L}_{\text{match}}
  &= \lambda_{\text{cls}}\,\mathfrak{L}_{\text{cls}}
  + \lambda_{\text{mask}}\,\mathfrak{L}_{\text{mask}}.
\end{split}
\]

\paragraph{Classification Loss (Focal Loss)}
For each matched prediction $\hat{\mathbf{y}}_i$, we compute
\[
  \mathfrak{L}_{\text{cls}}(\hat{\mathbf{y}}_i,\mathbf{y}_{\text{cls}})
  = -\sum_{c\in\{0,1\}} \alpha_c\,(1 - p_{i,c})^{\gamma}\,y_c\,\log(p_{i,c}),
\]
with $\alpha_c=1$ and $\gamma=1$ by default.

\paragraph{Mask Loss}
The mask loss combines Dice loss and binary mask focal loss:
\[
  \mathfrak{L}_{\text{mask}}
  = \mathfrak{L}_{\text{dice}} + \mathfrak{L}_{\text{bfocal}}.
\]

\paragraph{Dice Loss}
\[
  \mathfrak{L}_{\text{dice}}(\mathbf{m}_i,\mathbf{y}_{\text{mask}})
  = 1 - \frac{2\sum_{n=1}^N p_i^n\,g^n}
               {\sum_{n=1}^N p_i^n + \sum_{n=1}^N g^n}\,.
\]

\paragraph{Binary Mask Focal Loss}
\[
\begin{split}
  \mathfrak{L}_{\text{bfocal}}(\mathbf{m}_i,\mathbf{y}_{\text{mask}})
  &= -\frac{1}{N}\sum_{n=1}^N
    \Bigl[(1 - p_i^n)^{\gamma}g^n\log(p_i^n) \\
  &\quad + (p_i^n)^{\gamma}(1 - g^n)\log(1 - p_i^n)\Bigr],
\end{split}
\]
with $\gamma=1$.

\paragraph{Box Loss}
The box loss is the sum of an $L1$ loss and a Generalized IoU loss:
\[
  \mathfrak{L}_{\text{box}}
  = \mathfrak{L}_{L1} + \mathfrak{L}_{\text{giou}}.
\]

\paragraph{$L1$ Loss}
\[
  \mathfrak{L}_{L1}(\mathbf{b}_i,\mathbf{b}_{\text{gt}})
  = \bigl\lVert \mathbf{b}_i - \mathbf{b}_{\text{gt}}\bigr\rVert_1.
\]

\paragraph{Generalized IoU Loss}
First define
\[
\begin{split}
  \mathrm{IoU}(B_i,B_{\text{gt}})
  &= \frac{|B_i\cap B_{\text{gt}}|}{|B_i\cup B_{\text{gt}}|}, \\
  \mathrm{GIoU}(B_i,B_{\text{gt}})
  &= \mathrm{IoU}(B_i,B_{\text{gt}})
    - \frac{|C\setminus (B_i\cup B_{\text{gt}})|}{|C|},
\end{split}
\]
then
\[
  \mathfrak{L}_{\text{giou}}
  = 1 - \mathrm{GIoU}(B_i,B_{\text{gt}}).
\]

\paragraph{Overall Loss}
Only the matched prediction $\hat{\mathbf{y}}_{\text{pos}}$ contributes:
\[
  \mathfrak{L}
  = \underbrace{\lambda_{\text{cls}}\,\mathfrak{L}_{\text{cls}}
    + \lambda_{\text{mask}}\,\mathfrak{L}_{\text{mask}}}
    _{\mathfrak{L}_{\text{match}}}
    + \lambda_{\text{box}}\,\mathfrak{L}_{\text{box}}.
\]

\begin{figure*}[bp]
    \centering
    \includegraphics[width=0.95\linewidth]{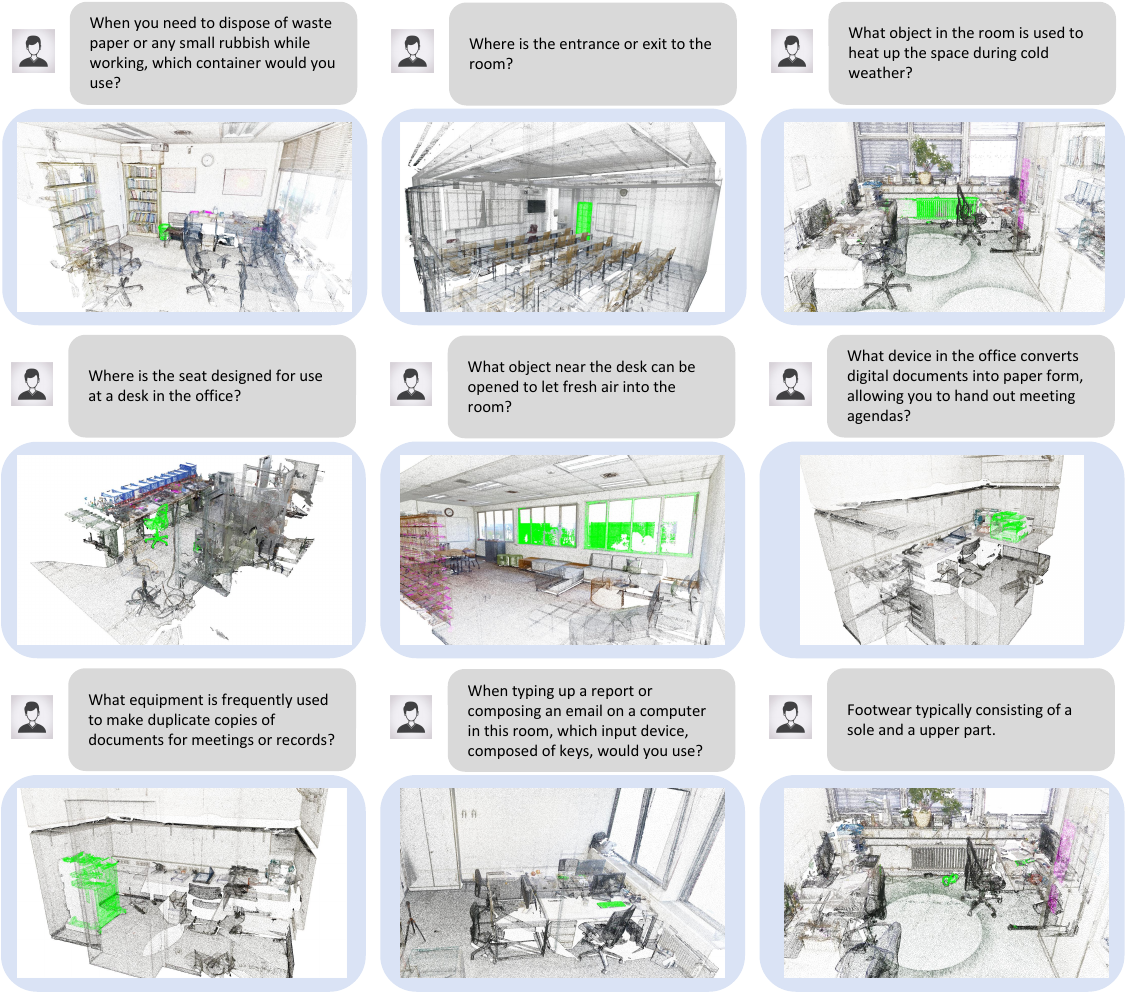}
    \caption{Visualization results on reasoning expression segmentation. We obtain those data from the partially open-source Instruc3D \cite{he2025segpoint} dataset. The segmented results are highlighted. Zoom in for better details.}
    \label{fig:reasonseg_vis}
\vspace{-5pt}
\end{figure*}

\begin{figure*}
    \centering
    \includegraphics[width=0.95\linewidth]{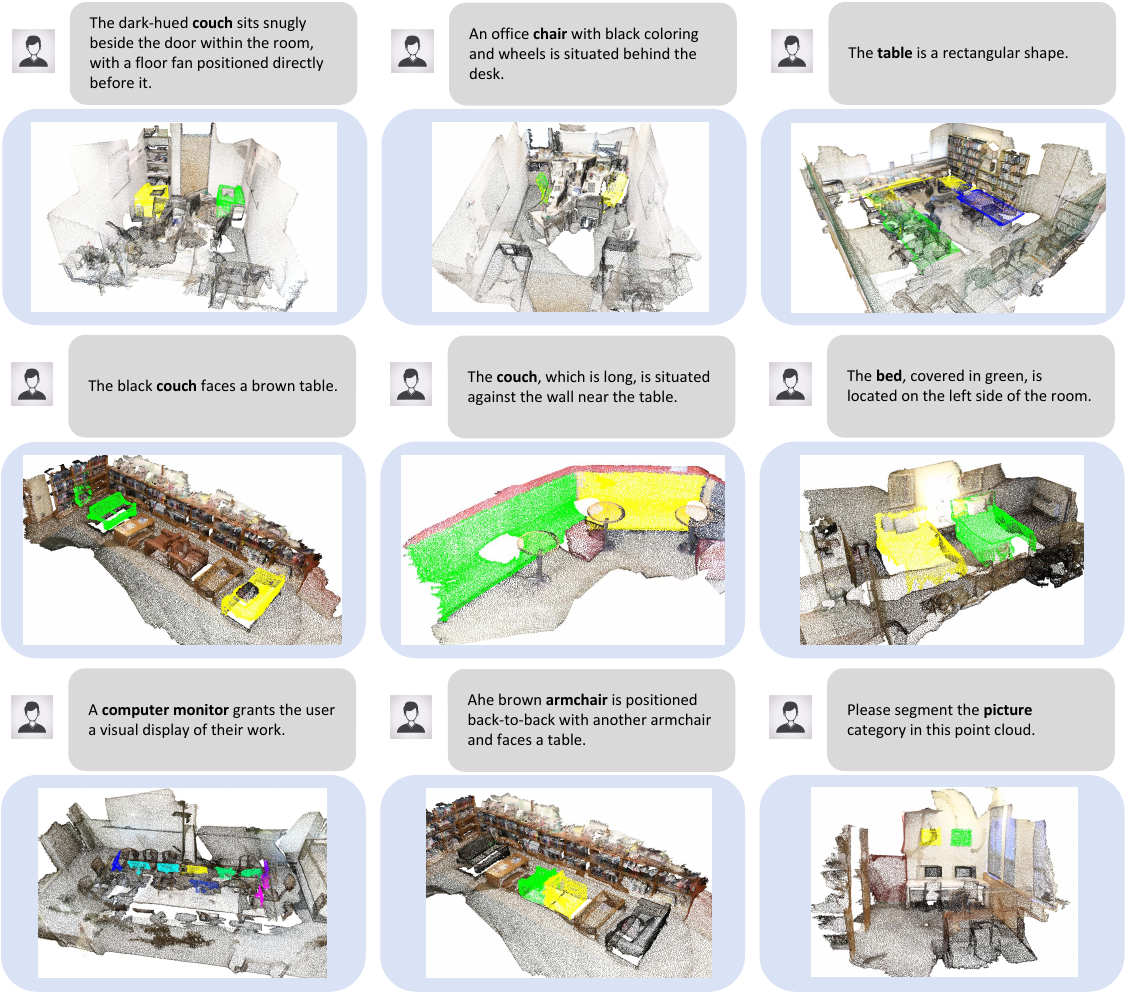}
    \caption{Visualization results on multi-object segmentation. Different instances are highlighted in different colors. Zoom-in for better details.}
    \label{fig:multiseg_vis}
\vspace{-5pt}
\end{figure*}

\subsection{Additional Visualization}
\label{sec:morevis}
The PLM is capable of reasoning implicit user instructions, even when the target object name is not explicitly provided in the language input, thereby enabling broader interactions through unrestricted commands. Figure \ref{fig:reasonseg_vis} shows the visualization results on different user commands.

Furthermore, PLM can easily handle complex scenes even when multiple objects are referred. Figure \ref{fig:multiseg_vis} presents multi-object segmentation scenarios, where PLM distinguishes multiple instances within a cluttered environment. PLM correctly identifies and separates objects that are referred by vague instructions, ensuring clear instance boundaries in its segmentations.

\subsection{Limitation Discussion}

The proposed OcDR and GRD are not limited to indoor scenes. In principle, our PLM can be applied to outdoor LiDAR data, CAD scenes, and settings with larger point counts. However, transfer to such domains is mainly constrained by the encoder rather than the PLM itself. Due to the substantial differences in scene scale, point density, and task formulation, effective transfer typically requires a well-aligned point–image–text encoder, which in turn depends on large-scale, high-quality multimodal paired data. Compared to indoor scenarios, such data remain relatively limited for outdoor LiDAR and CAD settings. 

% \newpage
\bibliographystyle{IEEEtran}
\bibliography{egbib}
% \vfill

\begin{IEEEbiography}[{\includegraphics[width=1in,height=1.25in,clip,keepaspectratio]{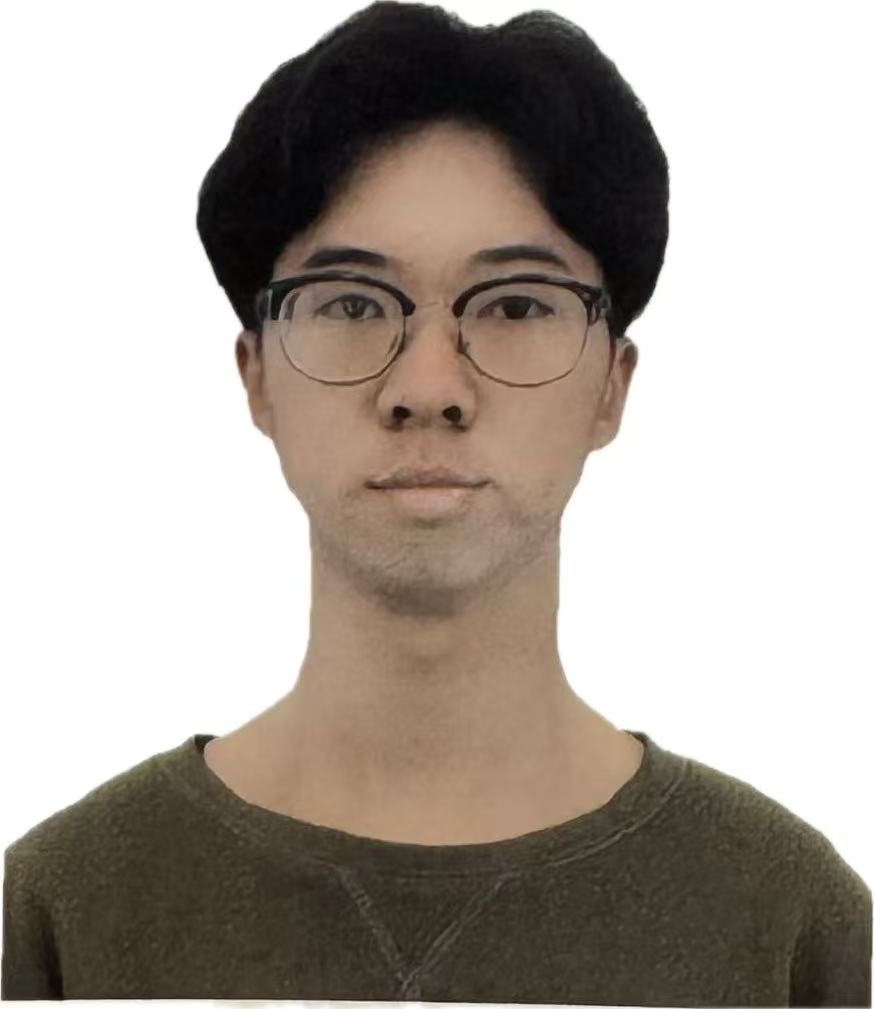}}]{Zhuoxu Huang} is a senior Ph.D. student in Computer Science at Aberystwyth University, UK. He received a bachelor’s degree from Wuhan University, China. He is a Vision Graphics and Visualisation Group member at Aberystwyth University. His research interests include 3D vision, video analysis, and multimodal LLM. 
\end{IEEEbiography}

\begin{IEEEbiography}[{\includegraphics[width=1in,height=1.25in,clip,keepaspectratio]{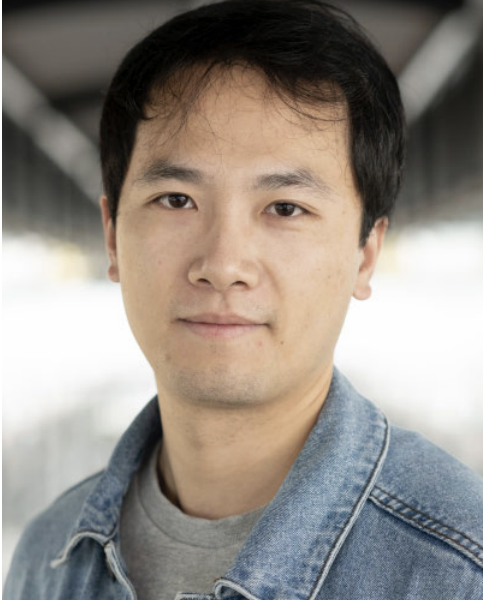}}]{Mingqi Gao} is a Postdoctoral Research Associate with the School of Computer Science, The University of Sheffield, U.K. He received the Ph.D. degree from the Warwick Manufacturing Group (WMG), University of Warwick, U.K., in 2024. His research interests include video foundation models, 3D vision, and multimodal analysis.
\end{IEEEbiography}

\begin{IEEEbiography}[{\includegraphics[width=1in,height=1.25in,clip,keepaspectratio]{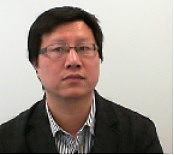}}]{Jungong Han} is a professor in the Department of Automation at Tsinghua University. He also holds an Honorary Professorship at the University of Warwick, UK. His research interests include computer vision, artificial intelligence, and machine learning. He is a Fellow of IAPR and a Fellow of AAIA. He serves as the Associate Editor for many prestigious journals, such as IEEE Transactions on Multimedia, IEEE Transactions on Image Processing, IEEE Transactions on Neural Networks and Learning Systems, IEEE Transactions on Circuits and Systems for Video Technology, and Pattern Recognition.
\end{IEEEbiography}

\end{document}